\pdfoutput=1

\documentclass[11pt]{article}

\usepackage[final]{EMNLP2023}

\usepackage{times}
\usepackage{latexsym}

\usepackage[T1]{fontenc}

\usepackage[utf8]{inputenc}

\usepackage{microtype}

\usepackage{inconsolata}

\usepackage[T1]{fontenc}    
\usepackage{hyperref}       
\usepackage{url}            
\usepackage{booktabs}       
\usepackage{amsfonts}       
\usepackage{nicefrac}       
\usepackage{microtype}      
\usepackage{xcolor}         
\usepackage{graphicx}
\usepackage{footnote}
\usepackage{listings}
\usepackage{lipsum}

\title{MusicAgent: An AI Agent for Music Understanding and Generation with Large Language Models}

\author{
Dingyao Yu$^{1,2}$, Kaitao Song$^{2}$, Peiling Lu$^2$, Tianyu He$^2$\\
\textbf{Xu Tan$^2$, Wei Ye$^{1*}$, Shikun Zhang$^{1*}$, Jiang Bian$^2$}\\
  Peking University$^1$, Microsoft Research Asia$^2$ \\ 
  \texttt{\{yudingyao, wye, zhangsk\}@pku.edu.cn},\\
  \texttt{\{kaitaosong, peil, tianyuhe, xuta, jiabia\}@microsoft.com}
\\ \\ 
\url{https://github.com/microsoft/muzic}
}

\begin{document}

\maketitle

\newcommand\blfootnote[1]{%
\begingroup
\renewcommand\thefootnote{}\footnote{#1}%
\addtocounter{footnote}{-1}%
\endgroup
}
\blfootnote{*Corresponding Author: Wei Ye, wye@pku.edu.cn; Shikun Zhang, zhangsk@pku.edu.cn}

\begin{abstract}

AI-empowered music processing is a diverse field that encompasses dozens of tasks, ranging from generation tasks (e.g., timbre synthesis) to comprehension tasks (e.g., music classification). For developers and amateurs, it is very difficult to grasp all of these task to satisfy their requirements in music processing, especially considering the huge differences in the representations of music data and the model applicability across platforms among various tasks. Consequently, it is necessary to build a system to organize and integrate these tasks, and thus help practitioners to automatically analyze their demand and call suitable tools as solutions to fulfill their requirements. 
Inspired by the recent success of large language models (LLMs) in task automation, we develop a system, named \textit{MusicAgent}, which integrates numerous music-related tools and an autonomous workflow to address user requirements. More specifically, we build 1) toolset that collects tools from diverse sources, including Hugging Face, GitHub, and Web API, etc. 2) an autonomous workflow empowered by LLMs (e.g., ChatGPT) to organize these tools and automatically decompose user requests into multiple sub-tasks and invoke corresponding music tools. The primary goal of this system is to free users from the intricacies of AI-music tools, enabling them to concentrate on the creative aspect. By granting users the freedom to effortlessly combine tools, the system offers a seamless and enriching music experience. 
The code is available on GitHub\footnote{\url{https://github.com/microsoft/muzic/tree/main/musicagent}} along with a brief instructional video\footnote{\url{https://youtu.be/tpNynjdcBqA}}.

\end{abstract}

\section{Introduction}

\begin{figure}[!t]
  \centering
  \includegraphics[width=\linewidth]{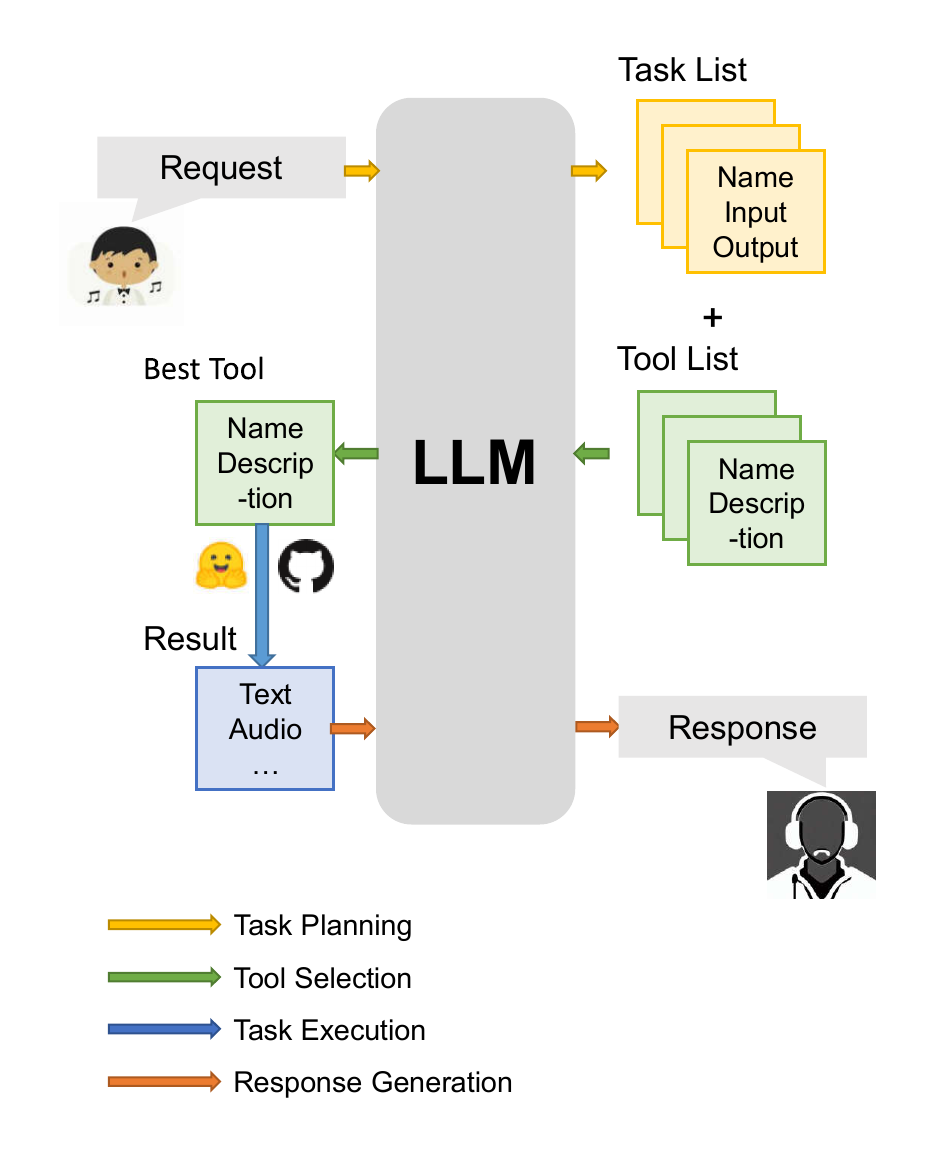}
  \caption{MusicAgent has gathered a rich collection of music-related tasks and diverse sources of tools, effectively integrating them with LLMs to achieve proficiency in handling complex music tasks.}
  \label{fig: intro}
\end{figure}

AI-empowered music processing is a multifaceted and intricate domain, encompassing a diverse range of tasks. Mastering this field presents a challenging endeavor due to the wide array of tasks it involves. 
Generally, the realm of music includes various generation and comprehension tasks, such as songwriting~\cite{sheng2021songmass, ju2021telemelody}, music generation~\cite{agostinelli2023musiclm, dai2021controllable, lu2023musecoco, lv2023getmusic}, audio transcription~\cite{benetos2018automatic, foscarin2020asap}, music retrieval~\cite{wu2023clamp}, etc. Specifically, music is a complex art form that weaves together a variety of distinct elements, such as chords and rhythm, to create vibrant and intricate content. 
Previous works have frequently encountered challenges in collaborating to complete complex music tasks, primarily due to differences in music feature designs and variations across platforms.
Therefore, how to build a system to automatically accomplish music-related tasks from the requests of users with varying levels of expertise is still an enticing direction worth exploring.


Recently, large language models (LLMs) have attracted considerable attention due to their outstanding performance in solving natural language processing (NLP) tasks~\cite{brown2020language, ouyang2022training, zhang2022opt, chowdhery2022palm, zeng2022glm, touvron2023llama}. 
The huge potentials of LLMs also inspire and directly facilitate many emerging techniques (e.g., in-context learning~\cite{xie2021explanation, min2022rethinking}, instruct tuning~\cite{longpre2023flan, wang2022super}, and chain-of-thought prompting~\cite{wei2022chain, kojima2022large}), which also further elevate the capability of LLMs. On the basis of these LLM capabilities, many researchers have extended the scope of LLMs to various topics. They borrow the idea of acting LLMs as the controllers to orchestrate various domain-specific expert models for solving complex AI tasks, such as HuggingGPT~\cite{shen2023hugginggpt}, AutoGPT and other modality-specifics ones~\cite{chen2022visualgpt, wu2023visual, huang2023audiogpt}. These successes also motivate us to explore the possibility to develop a system capable of assisting with various music-related tasks. 




Distinguishing from other modalities, incorporating LLMs with music presents the following features and challenges:
\begin{enumerate}
    \item \textbf{Tool Diversity}: On one hand, music-related tasks exhibit a wide range of diversity, and on the other hand, the corresponding tools for these tasks might not always reside on the same platform. These tools could be parameterized models available in open-source communities like GitHub, presented as software and applications, or even hosted through Web APIs for certain retrieval tasks. Considering all these factors is crucial when undertaking a comprehensive music workflow.
    \item \textbf{Cooperation}: The collaboration between music tools is also constrained by two factors. First, the diversity of music domain tasks leads to the absence of explicit input-output modality standards. Second, even when the modalities are identical, the music formats may differ, for instance, between symbolic music and audio music.
\end{enumerate}




To address these issues, we introduce MusicAgent, a specialized system designed to tackle the challenges. Inspired by recent work like HuggingGPT~\cite{shen2023hugginggpt}, MusicAgent is a framework that utilizes the power of LLMs as the controller and massive expert tools to accomplish user instructions, just as illustrated in Figure~\ref{fig: intro}. For the toolset, in addition to utilizing the models provided by Hugging Face, we further integrate various methods from different sources, including code from GitHub and Web APIs. To make collaboration between diverse tools, MusicAgent enforces standardized input-output formats across various tasks to promote seamless cooperation between tools. As a music-related system, all samples are trimmed to fit within a single audio segment, facilitating fundamental music operations among samples. For more system details and guidance on integrating additional tools, please refer to Section \ref{sec: method}.




\begin{figure*}[!t]
  \centering
  \includegraphics[width=0.9\textwidth]{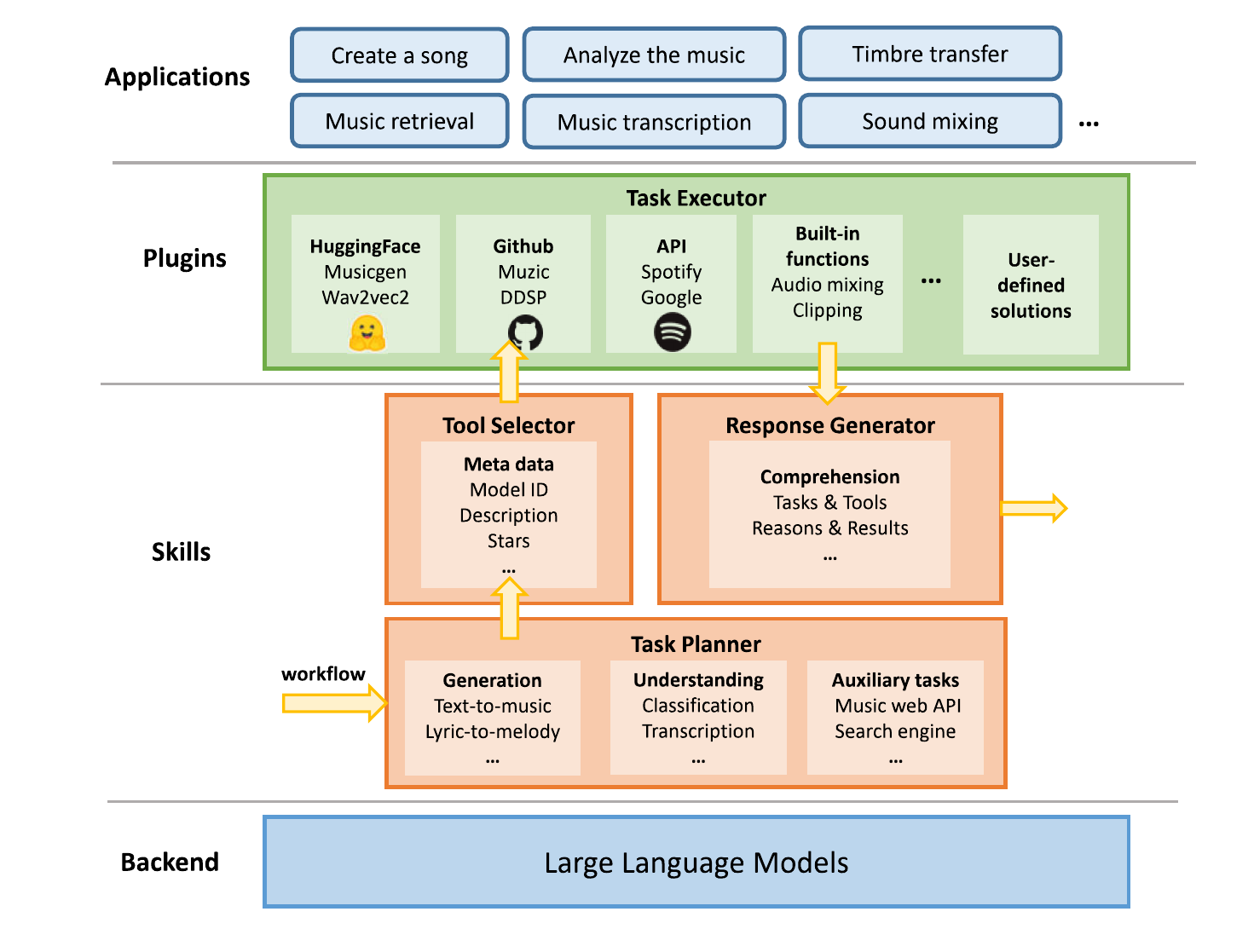}
  \caption{MusicAgent consists of four core components: the task planner, tool selector, task executor, and response generator. Among these, the task planner, tool selector, and response generator are built upon language language models (LLMs). When users make requests, MusicAgent decomposes and organizes the requests into subtasks. The system then selects the most suitable tool for each task. The chosen tool processes the input and populates the anticipated output. The LLM subsequently organizes the output, culminating in a comprehensive and efficient music processing system.}
  \label{fig: overview}
\end{figure*}

Overall, the MusicAgent presents several significant contributions:

\begin{itemize}
    \item \textbf{Accessibility:} MusicAgent eliminates the need to master complex AI music tools. By utilizing LLMs as the task planner, the system dynamically selects the most suitable methods for each music-related task, making music processing accessible to a broader audience.
    \item \textbf{Unity:} MusicAgent bridges the gap between tools from diverse sources by unifying the data format (e.g., text, MIDI, ABC notation, audio). The system enables seamless cooperation among tools on different platforms.
    \item \textbf{Modularity:} MusicAgent is highly extensible, allowing users to easily expand its functionality by implementing new functions, integrating GitHub projects, and incorporating Hugging Face models.
\end{itemize}


\section{Related Works}

\subsection{AI-Empowered Music Processing}

Music generation and understanding are multifaceted tasks that encompass various sub-tasks. In the realm of music generation, these tasks involve melody generation~\cite{yu2020lyrics, zhang2022relyme, yu2022museformer}, audio generation~\cite{donahue2018adversarial}, singing voice synthesis~\cite{ren2020deepsinger, lu2020xiaoicesing}, and sound mixing. In contrast, music understanding encompasses track separation~\cite{defossez2019demucs}, audio recognition, score transcription~\cite{2022_BittnerBRME_LightweightNoteTranscription_ICASSP}, audio classification~\cite{choi2017convolutional, zeng2021musicbert}, and music retrieval~\cite{wu2023clamp}. In addition to these diverse and complex music-related tasks, another significant challenge in traditional music processing is substantial differences in input and output formats across each task. These diversities in tasks and data formats also hinder the unification in music processing, which makes it difficult for us to develop a copilot for solving different musical tasks. Therefore, in this paper, we will discuss how to design a copilot to unified musical data format and combine these tools to automatically accomplish tasks by utilizing large language model.


\subsection{Large Language Models}
The field of natural language processing (NLP) is undergoing a revolutionary shift due to the emergence of large language models (LLMs). These models~\cite{brown2020language, touvron2023llama} have exhibited powerful performance in various language tasks, such as translation, dialogue modeling, and code completion, making them a focal point in NLP.


Based on these advantages, LLMs have been applied to many applications. Recently, a new trend is to use LLMs to build autonomous agents for task automation, just like AutoGPT~\footnote{\url{https://github.com/Significant-Gravitas/Auto-GPT}} and HuggingGPT~\cite{shen2023hugginggpt}. In these works, they will leverage an LLM as the controller to automatically analyze user requests and then invoke the appropriate tool for solving tasks. Although there are some successful trials in vision~\cite{chen2022visualgpt} or speech~\cite{huang2023audiogpt}, it is still challenging to build an autonomous agent for music processing, due to its diversity and complexity in tasks and data. Therefore, we present a system called MusicAgent, which integrates various functions to handle multiple music-related tasks, to accomplish requests from different users, including novices and professionals.



\begin{savenotes}
\begin{table*}
  \caption{Overview of tasks and the associated example tools in MusicAgent.}
  \label{tab: sample-table}
  \centering
  \begin{tabular}{l|ll|l|l}
    \toprule
    Task     & Input    & Output     & Task Type   & Example Tool \\
    \midrule
    text-to-symbolic-music & text   & symbolic music & Generation &  MuseCoco\footnote{\url{https://github.com/microsoft/muzic/tree/main/musecoco}}\\
    lyric-to-melody & text   & symbolic music & Generation &  ROC\footnote{\url{https://github.com/microsoft/muzic}}\\
    singing-voice-synthesis & text   & audio & Generation &  HiFiSinger\footnote{\url{https://github.com/CODEJIN/HiFiSinger}}\\
    text-to-audio & text   & audio & Generation &  AudioLDM\\
    timbre-transfer & audio   & audio & Generation &  DDSP\footnote{\url{https://github.com/magenta/ddsp}}\\
    accompaniment & symbolic music   & symbolic music & Generation &  GetMusic\footnote{\url{https://github.com/microsoft/muzic/tree/main/musecoco/getmusic}}\\
    music-classification & audio   & text & Understanding &  Wav2vec2\\
    music-separation & audio   & audio & Understanding &  Demucs\\
    lyric-recognition & audio   & text & Understanding &  Whisper-large-zh\footnote{\url{https://huggingface.co/jonatasgrosman/whisper-large-zh-cv11}}\\
    score-transcription & audio   & text & Understanding &  Basic-pitch\\
    artist/track-search & text   & audio & Auxiliary &  Spotify API\footnote{\url{https://spotify.com}}\\
    lyric-generation & text   & text & Auxiliary &  ChatGPT\\
    web-search & text   & text & Auxiliary &  Google API\\
    \bottomrule
  \end{tabular}
\end{table*}
\end{savenotes}
\section{MusicAgent}

\label{sec: method}


MusicAgent is a comprehensive system that enhances the capabilities of large language models (LLMs) and tailors them to the music domain by integrating additional data sources, dependent tools, and task specialization. As illustrated in Figure \ref{fig: overview}, MusicAgent designs an LLM-empowered autonomous workflow, which includes three key skills: Task Planner, Tool Selector, and Response Generator. These skills, along with the music-related tools forming the Task Executor, are integrated, resulting in a versatile system capable of executing various applications. In this section, we will delve into different aspects of this system, exploring its functionalities and contributions to the field of music processing.

\begin{figure}[!t]
  \centering
  \includegraphics[width=\linewidth]{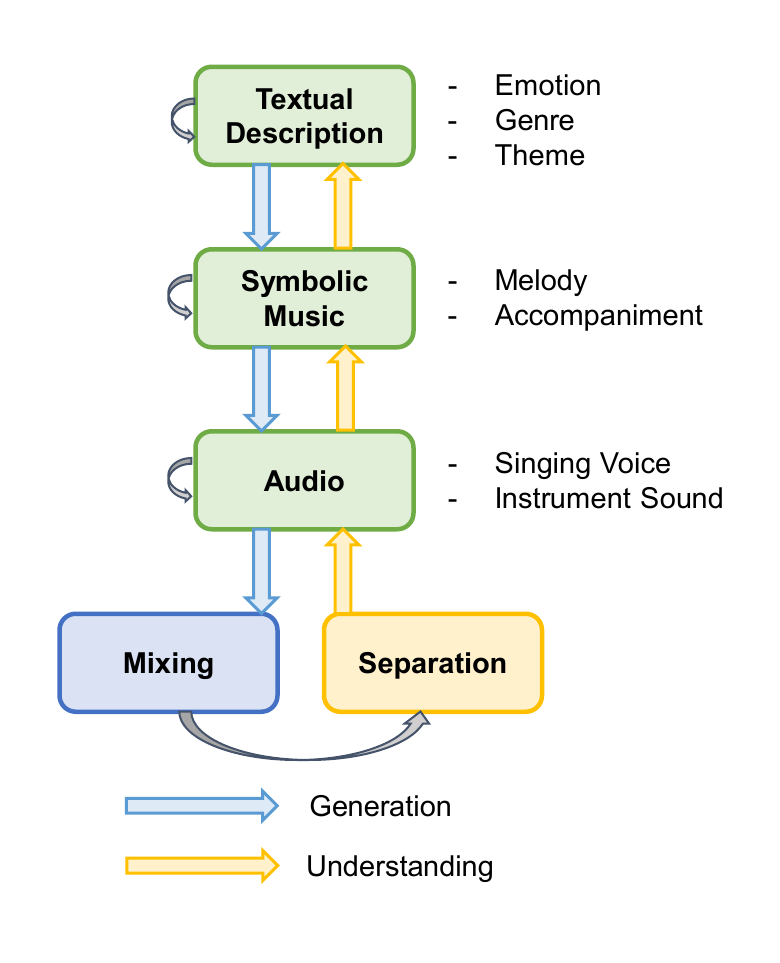}
  \caption{MusicAgent collects tasks and tools within the framework of music generation and understanding. It encompasses various tasks, including single-modal tasks and modality transfer tasks, such as converting sheet music to audio through singing voice synthesis.}
  \label{fig: framework}
\end{figure}

\begin{figure*}[!t]
  \centering
  \includegraphics[width=\textwidth]{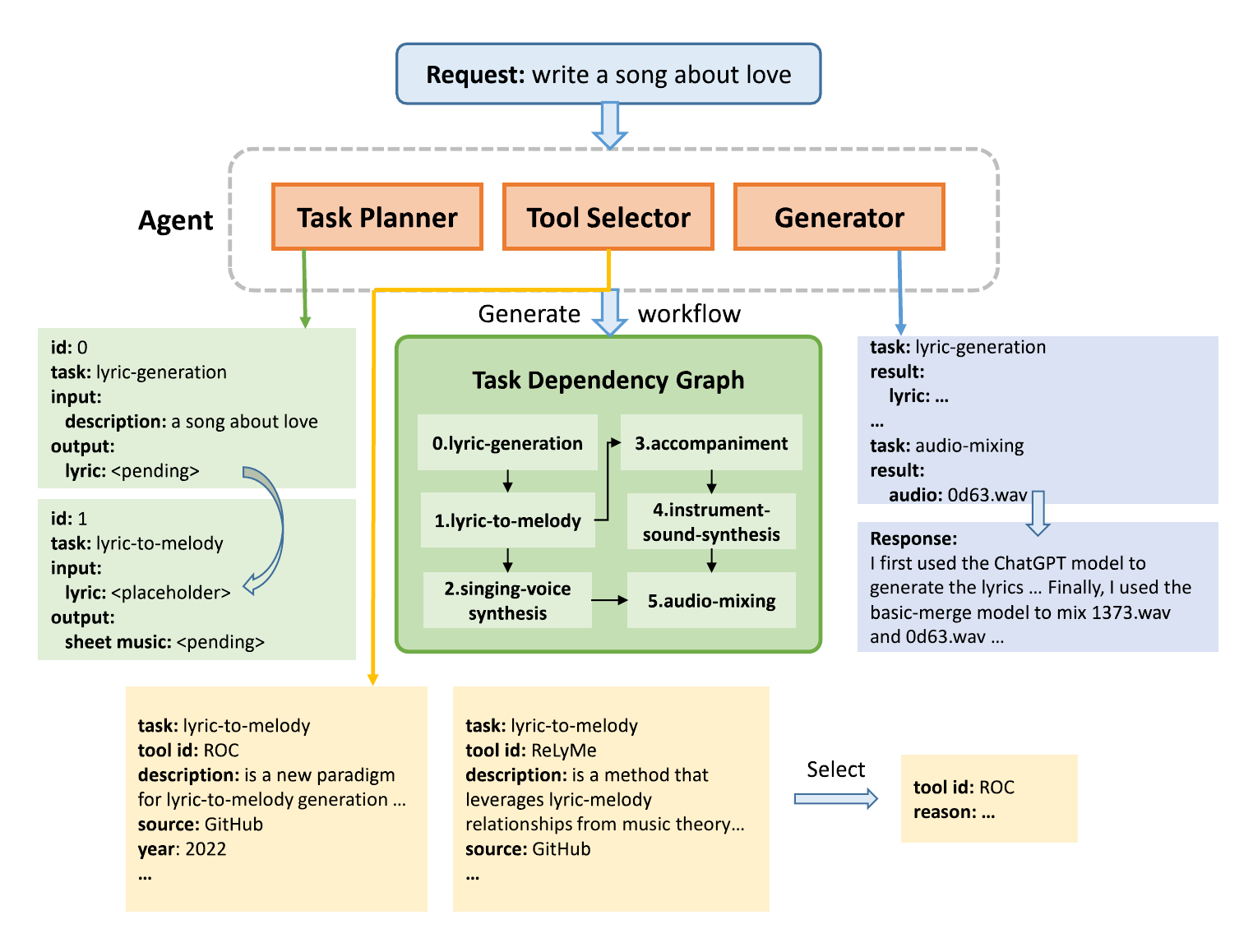}
  \caption{The LLM backend is responsible for the following steps: The Task Planner takes user requests and produces parsed task queue, the Tool Selector chooses suitable tools, and the Response Generator collects tool outputs and organizes the responses.}
  \label{fig: workflow}
\end{figure*}

\subsection{Tasks and Tools Collection}

Table \ref{tab: sample-table} provides a comprehensive overview of the music-related tasks and representative tools gathered in the current MusicAgent. We have organized the task sets based on the music processing flow illustrated in Figure \ref{fig: framework}. Aside from generation and understanding tasks, the collected tasks are primarily categorized into three groups:

\noindent\textbf{Generation tasks:} This category includes \textit{text-to-music}, \textit{lyric-to-melody}, \textit{singing-voice-synthesis}, \textit{timbre-transfer}, \textit{accompaniment}, and etc. 
 These tasks enable the collaborative music generation starting from simple descriptions.

\noindent\textbf{Understanding tasks:} The tasks of \textit{music-classification}, \textit{music-separation}, \textit{lyric recognition}, and \textit{music-transcription} are under this category. Combining these tasks enables the conversion of music into symbolic representation and the analysis of various music features.

\noindent\textbf{Auxiliary tasks:} This category encompasses web search and various audio processing toolkits. Web search includes text search using the \textit{Google API}, as well as music search through the \textit{Spotify API}. These tasks primarily provide rich data sources and perform basic operations on audio/MIDI/text data, serving as auxiliary functions.

Furthermore, Figure \ref{fig: framework} illustrates the utilization of three main data formats in the system: i) text, which includes lyric, genre or any other attributes related to the music. ii) sheet music, represented as MIDI files, describes the score of the music. iii) audio, containing the sound of the music.

\subsection{Autonomous Workflow}

The MusicAgent system consists of two parts: the autonomous workflow  and the plugins. The autonomous workflow serves as the core LLM interaction component, as shown in Figure~\ref{fig: overview}, and it comprises three skills: Task Planner, Tool Selector, and Response Generator, all supported by the LLM. Figure~\ref{fig: workflow} further demonstrates how these components work together harmoniously.

\noindent\textbf{Task Planner:} The Task Planner plays a critical role in converting user instructions into structured information, as most existing music tools only accept specialized inputs. 
The user input processed by the Task Planner will form the backbone of the entire workflow, encompassing the determination of each subtask and its corresponding input-output format, as well as the dependencies between the subtasks, creating a dependency graph.
Leveraging in-context learning, MusicAgent demonstrates excellent task decomposition performance. We provide task planner descriptions, supported tasks, and information structure in the prompt, along with several examples of music task-related decompositions. The user's interaction history and current input will replace the content at the corresponding position in the prompt. By utilizing the Semantic Kernel~\cite{Microsoft2023SK}, users can insert the required task flow in text format, thereby enhancing task planning effectiveness.

\noindent\textbf{Tool Selector:} The Tool Selector chooses the most appropriate tool from the open-source tools relevant to a specific subtask. Each tool is associated with its unique attributes, such as textual descriptions, download count, star ratings, and more. By incorporating these tool attributes with the user input, LLM presents the tool's ID and corresponding reasoning for what it deems the most suitable selection. Users have the flexibility to adjust the tool attributes and determine how LLM interprets these attributes. For instance, users can emphasize download count to meet diverse requirements.

\noindent\textbf{Response Generator:} The Response Generator gathers all intermediate results from the execution of subtasks and ultimately compiles them into a coherent response. Examples in Figure \ref{fig: conversation} demonstrate how LLM organizes the tasks and results to generate answers.




\subsection{Plugins}

When all the dependent tasks of a subtask have been completed, and all inputs have been instantiated, the LLM backend passes the task to the Task Executor, where the tool selects the necessary parameters from the inputs. Additionally, the tool needs to identify the task type, as a tool may handle multiple tasks.

MusicAgent stores model parameters on the CPU and only loads them into the GPU when actively in use. This approach is especially advantageous for users with limited GPU memory, as it optimizes resource utilization and ensures smooth task execution without overburdening the GPU memory.

\section{System Usage}

In this section, we will provide comprehensive guidelines on how to effectively use the MusicAgent toolkit.

\subsection{Code Usage}

Users have the flexibility to run this system either by following the instructions on GitHub or by integrating it as a module in their code or using it through the command line for more advanced usage, enabling the incorporation of custom tools. As depicted in Listing \ref{list: usage}, users can add custom task types, update tool attributes, and design prompts for each subtask, enhancing support for specific tasks. It is important to note that embedding the prompt in the history is a temporary operation, and there is a possibility of overlap if the context exceeds the limit. For permanent storage, it is recommended to directly include the prompt in the code.

\begin{lstlisting}[caption=Code usage of MusicAgent, label=list: usage]
## 1. Initialze the agent
from agent import MusicAgent
music_agent = MusicAgent(CONFIG_PATH)

## 2. Add custom tasks and tools
music_agent.task_map[MY_TASK].append(MY_TOOL)
music_agent.pipelines.append(MY_TOOL_CLASS)
# Update prompts
music_agent._init_task_context()
music_agent._init_tool_context()

## 3. Update tool's information
music_agent.update_tool_attributes(
    MY_TOOL, {"stars":..,"likes":..})
music_agent._init_tool_context()

## 4. Update the prompt
# Take task planner as an example
# There is a risk of being overwritten
music_agent.task_context["history"] +=
    "MY CUSTOM PROMPT"
    
## 5. Chat with the agent
music_agent.chat("Generate a song...")
\end{lstlisting}

\subsection{Demo Usage}

Apart from command-line usage, we have also provided a Gradio demo for users, where an OpenAI token is required. In the Gradio demo, users can directly upload audio and visually observe all the intermediate results generated by the system, as depicted in Figure \ref{fig: gradio}. Additionally, although MusicAgent includes built-in context truncation, users can still clear all LLM interaction history in the interface to refresh the agent.

\section{Conclusion}

In this paper, we introduce MusicAgent, an LLM-powered autonomous agent in the music domain. Our system can be considered as an auxiliary tool to help developers or audiences to automatically analyze user requests and select appropriate tools as solutions. Moreover, our framework directly integrates numerous music-related tools from various sources (e.g., Hugging Face, GitHub, Web search and etc). We also adapt the autonomous workflow to enable better compatibility in musical tasks and allow users to extend its toolset. In the future, we also further envision integrating more music-related functions into MusicAgent.



\section*{Acknowledgements}

We extend our gratitude to all anonymous reviewers and members of the Machine Learning group at Microsoft Research Asia for their valuable contributions and insightful suggestions in the development of this system.

\bibliography{emnlp2023}

\begin{thebibliography}{36}
\expandafter\ifx\csname natexlab\endcsname\relax\def\natexlab#1{#1}\fi

\bibitem[{Agostinelli et~al.(2023)Agostinelli, Denk, Borsos, Engel, Verzetti,
  Caillon, Huang, Jansen, Roberts, Tagliasacchi
  et~al.}]{agostinelli2023musiclm}
Andrea Agostinelli, Timo~I Denk, Zal{\'a}n Borsos, Jesse Engel, Mauro Verzetti,
  Antoine Caillon, Qingqing Huang, Aren Jansen, Adam Roberts, Marco
  Tagliasacchi, et~al. 2023.
\newblock Musiclm: Generating music from text.
\newblock \emph{arXiv preprint arXiv:2301.11325}.

\bibitem[{Benetos et~al.(2018)Benetos, Dixon, Duan, and
  Ewert}]{benetos2018automatic}
Emmanouil Benetos, Simon Dixon, Zhiyao Duan, and Sebastian Ewert. 2018.
\newblock Automatic music transcription: An overview.
\newblock \emph{IEEE Signal Processing Magazine}, 36(1):20--30.

\bibitem[{Bittner et~al.(2022)Bittner, Bosch, Rubinstein, Meseguer-Brocal, and
  Ewert}]{2022_BittnerBRME_LightweightNoteTranscription_ICASSP}
Rachel~M. Bittner, Juan~Jos\'e Bosch, David Rubinstein, Gabriel
  Meseguer-Brocal, and Sebastian Ewert. 2022.
\newblock A lightweight instrument-agnostic model for polyphonic note
  transcription and multipitch estimation.
\newblock In \emph{Proceedings of the IEEE International Conference on
  Acoustics, Speech, and Signal Processing (ICASSP)}, Singapore.

\bibitem[{Brown et~al.(2020)Brown, Mann, Ryder, Subbiah, Kaplan, Dhariwal,
  Neelakantan, Shyam, Sastry, Askell et~al.}]{brown2020language}
Tom Brown, Benjamin Mann, Nick Ryder, Melanie Subbiah, Jared~D Kaplan, Prafulla
  Dhariwal, Arvind Neelakantan, Pranav Shyam, Girish Sastry, Amanda Askell,
  et~al. 2020.
\newblock Language models are few-shot learners.
\newblock \emph{Advances in neural information processing systems},
  33:1877--1901.

\bibitem[{Chen et~al.(2022)Chen, Guo, Yi, Li, and
  Elhoseiny}]{chen2022visualgpt}
Jun Chen, Han Guo, Kai Yi, Boyang Li, and Mohamed Elhoseiny. 2022.
\newblock Visualgpt: Data-efficient adaptation of pretrained language models
  for image captioning.
\newblock In \emph{Proceedings of the IEEE/CVF Conference on Computer Vision
  and Pattern Recognition}, pages 18030--18040.

\bibitem[{Choi et~al.(2017)Choi, Fazekas, Sandler, and
  Cho}]{choi2017convolutional}
Keunwoo Choi, Gy{\"o}rgy Fazekas, Mark Sandler, and Kyunghyun Cho. 2017.
\newblock Convolutional recurrent neural networks for music classification.
\newblock In \emph{2017 IEEE International conference on acoustics, speech and
  signal processing (ICASSP)}, pages 2392--2396. IEEE.

\bibitem[{Chowdhery et~al.(2022)Chowdhery, Narang, Devlin, Bosma, Mishra,
  Roberts, Barham, Chung, Sutton, Gehrmann et~al.}]{chowdhery2022palm}
Aakanksha Chowdhery, Sharan Narang, Jacob Devlin, Maarten Bosma, Gaurav Mishra,
  Adam Roberts, Paul Barham, Hyung~Won Chung, Charles Sutton, Sebastian
  Gehrmann, et~al. 2022.
\newblock Palm: Scaling language modeling with pathways.
\newblock \emph{arXiv preprint arXiv:2204.02311}.

\bibitem[{Dai et~al.(2021)Dai, Jin, Gomes, and
  Dannenberg}]{dai2021controllable}
Shuqi Dai, Zeyu Jin, Celso Gomes, and Roger~B Dannenberg. 2021.
\newblock Controllable deep melody generation via hierarchical music structure
  representation.
\newblock \emph{arXiv preprint arXiv:2109.00663}.

\bibitem[{D{\'e}fossez et~al.(2019)D{\'e}fossez, Usunier, Bottou, and
  Bach}]{defossez2019demucs}
Alexandre D{\'e}fossez, Nicolas Usunier, L{\'e}on Bottou, and Francis Bach.
  2019.
\newblock Demucs: Deep extractor for music sources with extra unlabeled data
  remixed.
\newblock \emph{arXiv preprint arXiv:1909.01174}.

\bibitem[{Donahue et~al.(2018)Donahue, McAuley, and
  Puckette}]{donahue2018adversarial}
Chris Donahue, Julian McAuley, and Miller Puckette. 2018.
\newblock Adversarial audio synthesis.
\newblock \emph{arXiv preprint arXiv:1802.04208}.

\bibitem[{Foscarin et~al.(2020)Foscarin, Mcleod, Rigaux, Jacquemard, and
  Sakai}]{foscarin2020asap}
Francesco Foscarin, Andrew Mcleod, Philippe Rigaux, Florent Jacquemard, and
  Masahiko Sakai. 2020.
\newblock Asap: a dataset of aligned scores and performances for piano
  transcription.
\newblock In \emph{International Society for Music Information Retrieval
  Conference}, CONF, pages 534--541.

\bibitem[{Huang et~al.(2023)Huang, Li, Yang, Shi, Chang, Ye, Wu, Hong, Huang,
  Liu et~al.}]{huang2023audiogpt}
Rongjie Huang, Mingze Li, Dongchao Yang, Jiatong Shi, Xuankai Chang, Zhenhui
  Ye, Yuning Wu, Zhiqing Hong, Jiawei Huang, Jinglin Liu, et~al. 2023.
\newblock Audiogpt: Understanding and generating speech, music, sound, and
  talking head.
\newblock \emph{arXiv preprint arXiv:2304.12995}.

\bibitem[{Ju et~al.(2021)Ju, Lu, Tan, Wang, Zhang, Wu, Zhang, Li, Qin, and
  Liu}]{ju2021telemelody}
Zeqian Ju, Peiling Lu, Xu~Tan, Rui Wang, Chen Zhang, Songruoyao Wu, Kejun
  Zhang, Xiangyang Li, Tao Qin, and Tie-Yan Liu. 2021.
\newblock Telemelody: Lyric-to-melody generation with a template-based
  two-stage method.
\newblock \emph{arXiv preprint arXiv:2109.09617}.

\bibitem[{Kojima et~al.(2022)Kojima, Gu, Reid, Matsuo, and
  Iwasawa}]{kojima2022large}
Takeshi Kojima, Shixiang~Shane Gu, Machel Reid, Yutaka Matsuo, and Yusuke
  Iwasawa. 2022.
\newblock Large language models are zero-shot reasoners.
\newblock \emph{Advances in neural information processing systems},
  35:22199--22213.

\bibitem[{Longpre et~al.(2023)Longpre, Hou, Vu, Webson, Chung, Tay, Zhou, Le,
  Zoph, Wei et~al.}]{longpre2023flan}
Shayne Longpre, Le~Hou, Tu~Vu, Albert Webson, Hyung~Won Chung, Yi~Tay, Denny
  Zhou, Quoc~V Le, Barret Zoph, Jason Wei, et~al. 2023.
\newblock The flan collection: Designing data and methods for effective
  instruction tuning.
\newblock \emph{arXiv preprint arXiv:2301.13688}.

\bibitem[{Lu et~al.(2020)Lu, Wu, Luan, Tan, and Zhou}]{lu2020xiaoicesing}
Peiling Lu, Jie Wu, Jian Luan, Xu~Tan, and Li~Zhou. 2020.
\newblock Xiaoicesing: A high-quality and integrated singing voice synthesis
  system.
\newblock \emph{arXiv preprint arXiv:2006.06261}.

\bibitem[{Lu et~al.(2023)Lu, Xu, Kang, Yu, Xing, Tan, and
  Bian}]{lu2023musecoco}
Peiling Lu, Xin Xu, Chenfei Kang, Botao Yu, Chengyi Xing, Xu~Tan, and Jiang
  Bian. 2023.
\newblock Musecoco: Generating symbolic music from text.
\newblock \emph{arXiv preprint arXiv:2306.00110}.

\bibitem[{Lv et~al.(2023)Lv, Tan, Lu, Ye, Zhang, Bian, and
  Yan}]{lv2023getmusic}
Ang Lv, Xu~Tan, Peiling Lu, Wei Ye, Shikun Zhang, Jiang Bian, and Rui Yan.
  2023.
\newblock Getmusic: Generating any music tracks with a unified representation
  and diffusion framework.
\newblock \emph{arXiv preprint arXiv:2305.10841}.

\bibitem[{Microsoft(2023)}]{Microsoft2023SK}
Microsoft. 2023.
\newblock Semantic kernel.
\newblock \url{https://github.com/microsoft/semantic-kernel}.

\bibitem[{Min et~al.(2022)Min, Lyu, Holtzman, Artetxe, Lewis, Hajishirzi, and
  Zettlemoyer}]{min2022rethinking}
Sewon Min, Xinxi Lyu, Ari Holtzman, Mikel Artetxe, Mike Lewis, Hannaneh
  Hajishirzi, and Luke Zettlemoyer. 2022.
\newblock Rethinking the role of demonstrations: What makes in-context learning
  work?
\newblock \emph{arXiv preprint arXiv:2202.12837}.

\bibitem[{Ouyang et~al.(2022)Ouyang, Wu, Jiang, Almeida, Wainwright, Mishkin,
  Zhang, Agarwal, Slama, Ray et~al.}]{ouyang2022training}
Long Ouyang, Jeffrey Wu, Xu~Jiang, Diogo Almeida, Carroll Wainwright, Pamela
  Mishkin, Chong Zhang, Sandhini Agarwal, Katarina Slama, Alex Ray, et~al.
  2022.
\newblock Training language models to follow instructions with human feedback.
\newblock \emph{Advances in Neural Information Processing Systems},
  35:27730--27744.

\bibitem[{Ren et~al.(2020)Ren, Tan, Qin, Luan, Zhao, and
  Liu}]{ren2020deepsinger}
Yi~Ren, Xu~Tan, Tao Qin, Jian Luan, Zhou Zhao, and Tie-Yan Liu. 2020.
\newblock Deepsinger: Singing voice synthesis with data mined from the web.
\newblock In \emph{Proceedings of the 26th ACM SIGKDD International Conference
  on Knowledge Discovery \& Data Mining}, pages 1979--1989.

\bibitem[{Shen et~al.(2023)Shen, Song, Tan, Li, Lu, and
  Zhuang}]{shen2023hugginggpt}
Yongliang Shen, Kaitao Song, Xu~Tan, Dongsheng Li, Weiming Lu, and Yueting
  Zhuang. 2023.
\newblock Hugginggpt: Solving ai tasks with chatgpt and its friends in
  huggingface.
\newblock \emph{arXiv preprint arXiv:2303.17580}.

\bibitem[{Sheng et~al.(2021)Sheng, Song, Tan, Ren, Ye, Zhang, and
  Qin}]{sheng2021songmass}
Zhonghao Sheng, Kaitao Song, Xu~Tan, Yi~Ren, Wei Ye, Shikun Zhang, and Tao Qin.
  2021.
\newblock Songmass: Automatic song writing with pre-training and alignment
  constraint.
\newblock In \emph{Proceedings of the AAAI Conference on Artificial
  Intelligence}, volume~35, pages 13798--13805.

\bibitem[{Touvron et~al.(2023)Touvron, Lavril, Izacard, Martinet, Lachaux,
  Lacroix, Rozi{\`e}re, Goyal, Hambro, Azhar et~al.}]{touvron2023llama}
Hugo Touvron, Thibaut Lavril, Gautier Izacard, Xavier Martinet, Marie-Anne
  Lachaux, Timoth{\'e}e Lacroix, Baptiste Rozi{\`e}re, Naman Goyal, Eric
  Hambro, Faisal Azhar, et~al. 2023.
\newblock Llama: Open and efficient foundation language models.
\newblock \emph{arXiv preprint arXiv:2302.13971}.

\bibitem[{Wang et~al.(2022)Wang, Mishra, Alipoormolabashi, Kordi, Mirzaei,
  Arunkumar, Ashok, Dhanasekaran, Naik, Stap et~al.}]{wang2022super}
Yizhong Wang, Swaroop Mishra, Pegah Alipoormolabashi, Yeganeh Kordi, Amirreza
  Mirzaei, Anjana Arunkumar, Arjun Ashok, Arut~Selvan Dhanasekaran, Atharva
  Naik, David Stap, et~al. 2022.
\newblock Super-naturalinstructions: Generalization via declarative
  instructions on 1600+ nlp tasks.
\newblock \emph{arXiv preprint arXiv:2204.07705}.

\bibitem[{Wei et~al.(2022)Wei, Wang, Schuurmans, Bosma, Xia, Chi, Le, Zhou
  et~al.}]{wei2022chain}
Jason Wei, Xuezhi Wang, Dale Schuurmans, Maarten Bosma, Fei Xia, Ed~Chi, Quoc~V
  Le, Denny Zhou, et~al. 2022.
\newblock Chain-of-thought prompting elicits reasoning in large language
  models.
\newblock \emph{Advances in Neural Information Processing Systems},
  35:24824--24837.

\bibitem[{Wu et~al.(2023{\natexlab{a}})Wu, Yin, Qi, Wang, Tang, and
  Duan}]{wu2023visual}
Chenfei Wu, Shengming Yin, Weizhen Qi, Xiaodong Wang, Zecheng Tang, and Nan
  Duan. 2023{\natexlab{a}}.
\newblock Visual chatgpt: Talking, drawing and editing with visual foundation
  models.
\newblock \emph{arXiv preprint arXiv:2303.04671}.

\bibitem[{Wu et~al.(2023{\natexlab{b}})Wu, Yu, Tan, and Sun}]{wu2023clamp}
Shangda Wu, Dingyao Yu, Xu~Tan, and Maosong Sun. 2023{\natexlab{b}}.
\newblock Clamp: Contrastive language-music pre-training for cross-modal
  symbolic music information retrieval.
\newblock \emph{arXiv preprint arXiv:2304.11029}.

\bibitem[{Xie et~al.(2021)Xie, Raghunathan, Liang, and Ma}]{xie2021explanation}
Sang~Michael Xie, Aditi Raghunathan, Percy Liang, and Tengyu Ma. 2021.
\newblock An explanation of in-context learning as implicit bayesian inference.
\newblock \emph{arXiv preprint arXiv:2111.02080}.

\bibitem[{Yu et~al.(2022)Yu, Lu, Wang, Hu, Tan, Ye, Zhang, Qin, and
  Liu}]{yu2022museformer}
Botao Yu, Peiling Lu, Rui Wang, Wei Hu, Xu~Tan, Wei Ye, Shikun Zhang, Tao Qin,
  and Tie-Yan Liu. 2022.
\newblock Museformer: Transformer with fine-and coarse-grained attention for
  music generation.
\newblock \emph{Advances in Neural Information Processing Systems},
  35:1376--1388.

\bibitem[{Yu et~al.(2020)Yu, Harsco{\"e}t, Canales, Reddy~M, Tang, and
  Jiang}]{yu2020lyrics}
Yi~Yu, Florian Harsco{\"e}t, Simon Canales, Gurunath Reddy~M, Suhua Tang, and
  Junjun Jiang. 2020.
\newblock Lyrics-conditioned neural melody generation.
\newblock In \emph{MultiMedia Modeling: 26th International Conference, MMM
  2020, Daejeon, South Korea, January 5--8, 2020, Proceedings, Part II 26},
  pages 709--714. Springer.

\bibitem[{Zeng et~al.(2022)Zeng, Liu, Du, Wang, Lai, Ding, Yang, Xu, Zheng, Xia
  et~al.}]{zeng2022glm}
Aohan Zeng, Xiao Liu, Zhengxiao Du, Zihan Wang, Hanyu Lai, Ming Ding, Zhuoyi
  Yang, Yifan Xu, Wendi Zheng, Xiao Xia, et~al. 2022.
\newblock Glm-130b: An open bilingual pre-trained model.
\newblock \emph{arXiv preprint arXiv:2210.02414}.

\bibitem[{Zeng et~al.(2021)Zeng, Tan, Wang, Ju, Qin, and
  Liu}]{zeng2021musicbert}
Mingliang Zeng, Xu~Tan, Rui Wang, Zeqian Ju, Tao Qin, and Tie-Yan Liu. 2021.
\newblock Musicbert: Symbolic music understanding with large-scale
  pre-training.
\newblock \emph{arXiv preprint arXiv:2106.05630}.

\bibitem[{Zhang et~al.(2022{\natexlab{a}})Zhang, Chang, Wu, Tan, Qin, Liu, and
  Zhang}]{zhang2022relyme}
Chen Zhang, Luchin Chang, Songruoyao Wu, Xu~Tan, Tao Qin, Tie-Yan Liu, and
  Kejun Zhang. 2022{\natexlab{a}}.
\newblock Relyme: Improving lyric-to-melody generation by incorporating
  lyric-melody relationships.
\newblock In \emph{Proceedings of the 30th ACM International Conference on
  Multimedia}, pages 1047--1056.

\bibitem[{Zhang et~al.(2022{\natexlab{b}})Zhang, Roller, Goyal, Artetxe, Chen,
  Chen, Dewan, Diab, Li, Lin et~al.}]{zhang2022opt}
Susan Zhang, Stephen Roller, Naman Goyal, Mikel Artetxe, Moya Chen, Shuohui
  Chen, Christopher Dewan, Mona Diab, Xian Li, Xi~Victoria Lin, et~al.
  2022{\natexlab{b}}.
\newblock Opt: Open pre-trained transformer language models.
\newblock \emph{arXiv preprint arXiv:2205.01068}.

\end{thebibliography}
\bibliographystyle{acl_natbib}

\appendix

\section{Appendix}
\label{sec:appendix}

\begin{figure}
  \centering
  \includegraphics[width=\linewidth]{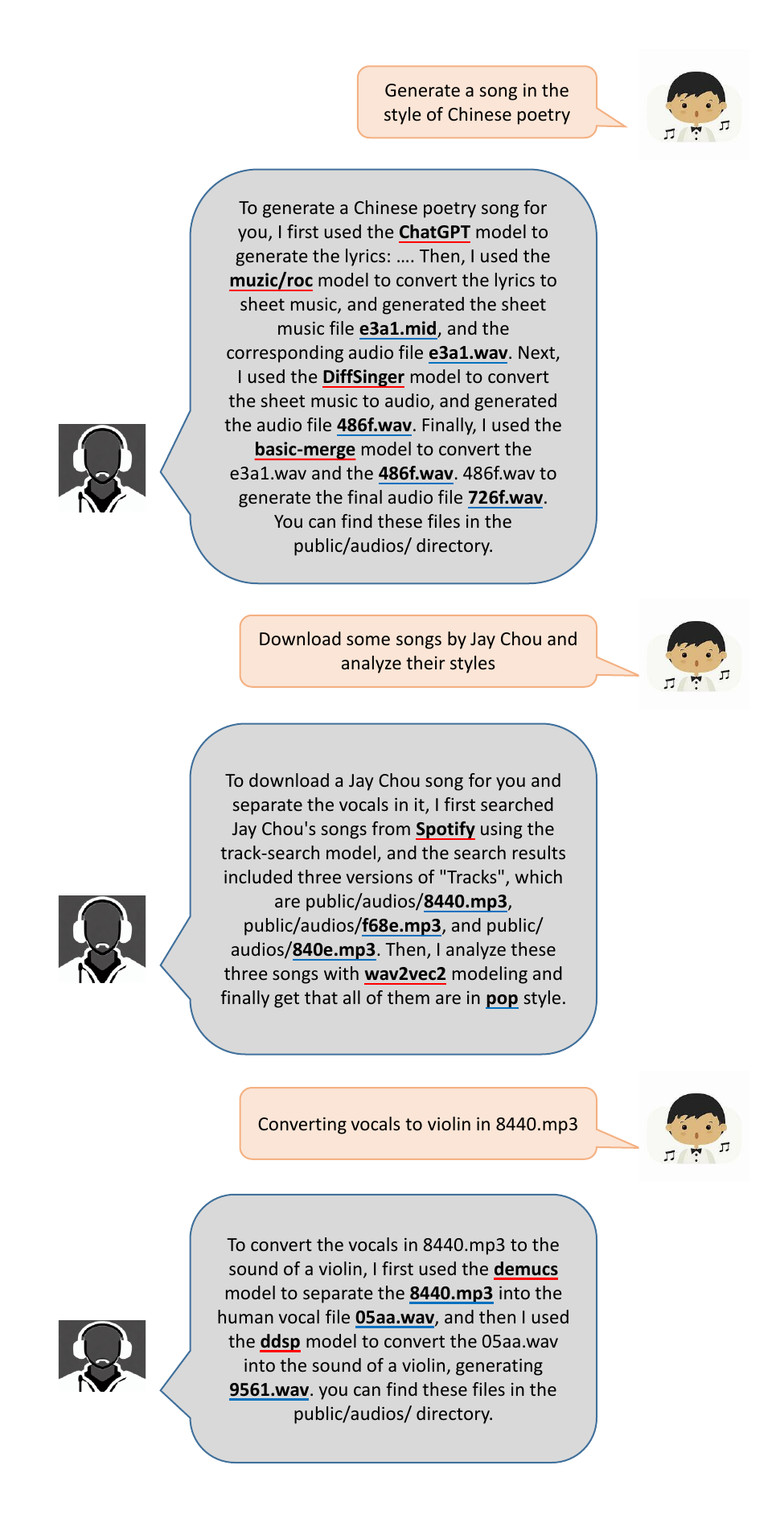}
  \caption{Sample Q\&A for users and agents.}
  \label{fig: conversation}
\end{figure}

\begin{figure*}[!t]
  \centering
  \includegraphics[width=\linewidth]{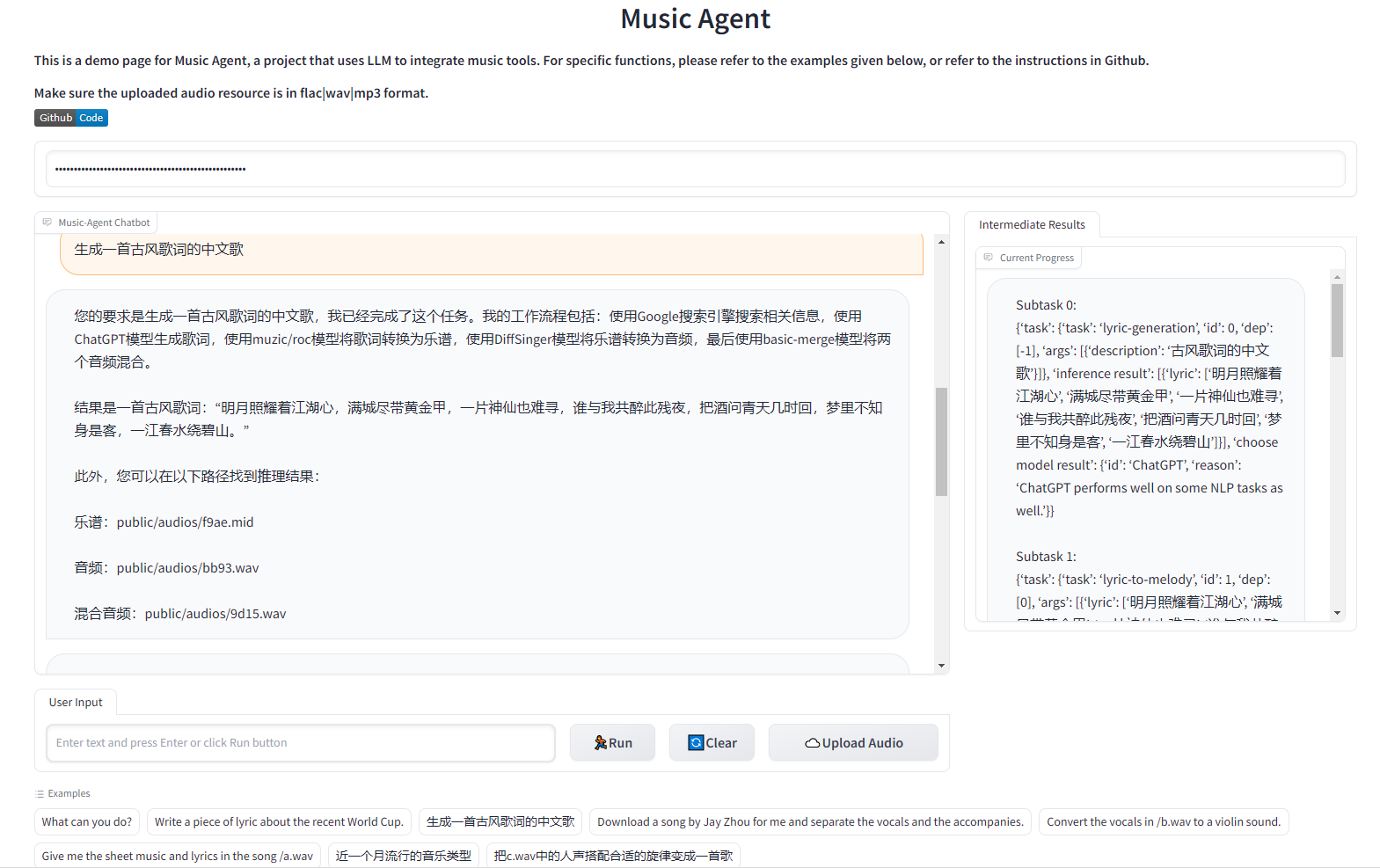}
  \caption{Gradio Demomstration.}
  \label{fig: gradio}
\end{figure*}

\end{document}